\documentclass{article}
\usepackage{spconf,amsmath,amsfonts,epsfig}
\usepackage{textcomp}
\usepackage{verbatim}
\usepackage{url}
\usepackage{xcolor}

\begin{document}\sloppy

\def\x{{\mathbf x}}
\def\L{{\cal L}}

\title{Regularizing Face Verification Nets For Pain Intensity Regression}
\name{\small Feng Wang$^{1,4}$, Xiang Xiang$^{1*}$\thanks{* Corresponding author (e-mail: \url{xxiang@cs.jhu.edu})}, Chang Liu$^{1,5}$, Trac D. Tran$^2$, Austin Reiter$^1$, Gregory D. Hager$^1$, Harry Quon$^3$, Jian Cheng$^4$, Alan L. Yuille$^1$ }

\address{
{Dept. of \{$^1$ Computer Science, $^2$ Electrical \& Computer Engineering, $^3$Radiation Oncology\}}\\
{Johns Hopkins University, 3400 N. Charles St, Baltimore, MD 21218, USA}\\
{$^4$ Dept. of EE, UESTC, 2006 Xiyuan Ave, Chengdu, Sichuan 611731, China}\\
{$^5$ Dept. of CS, Tsinghua University, Beijing 100084, China}
}

\maketitle

\begin{abstract}
Limited labeled data are available for the research of estimating facial expression intensities. 
For instance, the ability to train deep networks for automated pain assessment is limited by small datasets with labels of patient-reported pain intensities.
Fortunately, fine-tuning from a data-extensive pre-trained domain, such as face verification, can alleviate this problem. 
In this paper, we propose a network that fine-tunes a state-of-the-art face verification network using a regularized regression loss and additional data with expression labels. 
In this way, the expression intensity regression task can benefit from the rich feature representations trained on a huge amount of data for face verification.
The proposed regularized deep regressor is applied to estimate the pain expression intensity and verified on the widely-used UNBC-McMaster Shoulder-Pain dataset, achieving the state-of-the-art performance.
A weighted evaluation metric is also proposed to address the imbalance issue of different pain intensities.
\end{abstract}

\section{Introduction}

\begin{figure}[!t]
    \centering
    \includegraphics[scale=0.6]{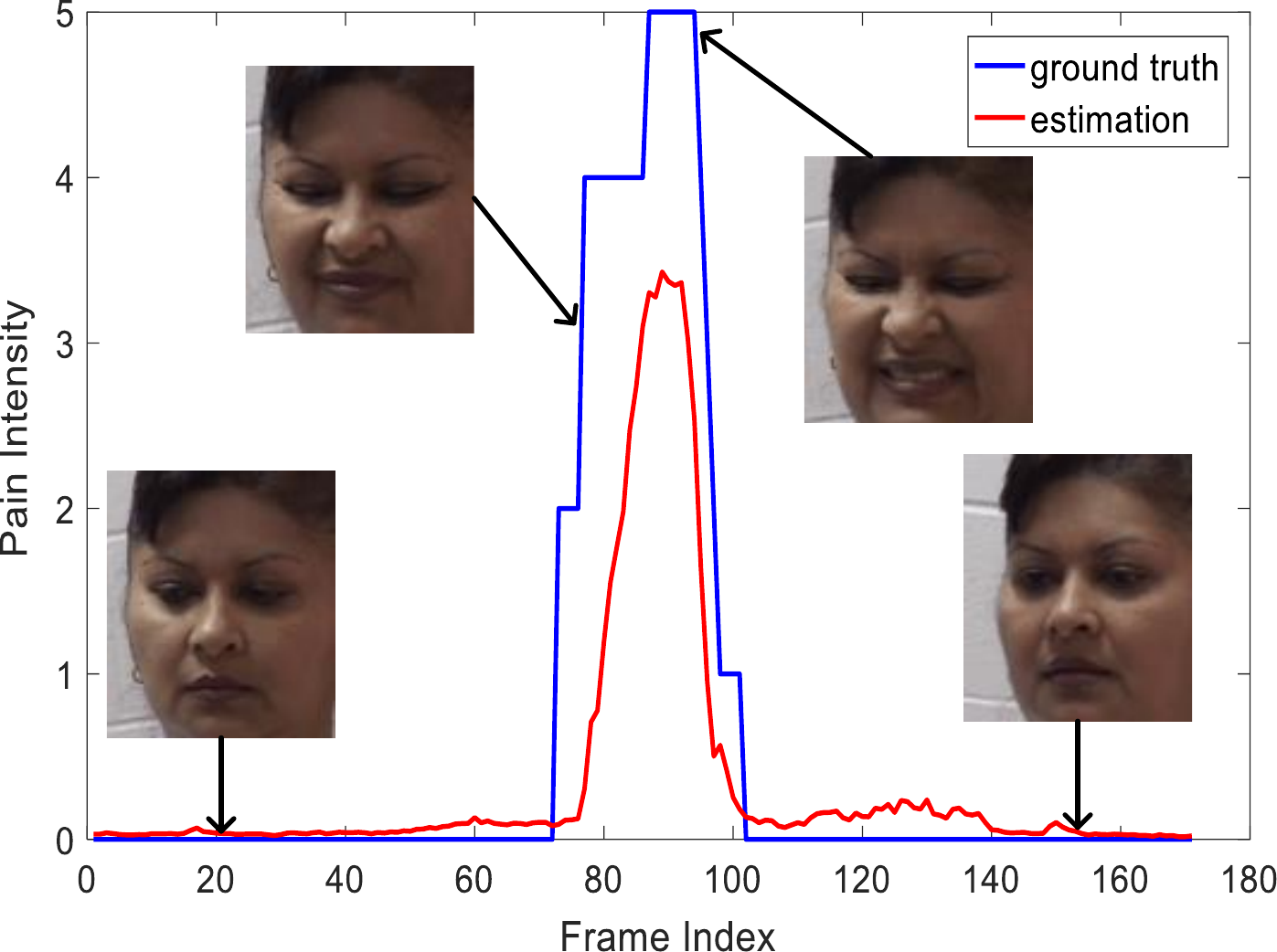}
    \caption{Example testing result of estimated pain intensities (see the continuous red curve) of one patient in one video from the Shoulder-Pain dataset \cite{lucey2011painful} which provides per-frame observer-rated labels (see the blue curve connected from discrete points of $(frame, intensity)$). Best viewed in color. }
    \label{fig:curve}
\end{figure}
\vspace{-2mm}

Obtaining accurate patient-reported pain intensities is important to effectively manage pain and thus reduce anesthetic doses and in-hospital deterioration.
Traditionally, caregivers work with patients to manually input the patients' pain intensity, 
ranging among a few levels such as mild, moderate, severe and excruciating.
Recently, a couple of concepts have been proposed such as active, automated and objective pain monitoring
over the patient's stay in hospital, with roughly the same motivation:
first to simplify the pain reporting process and reduce the strain on manual efforts;
second to standardize the feedback mechanism by ensuring a single metric that performs all assessments and thus reduces bias.
There indeed exist efforts to assess pain from the observational or behavioral effect caused by pain such as physiological data. \textcopyright Medasense has developed medical devices for objective pain monitoring. 
Their basic premise is that pain may cause vital signs such as blood pressure, pulse rate, respiration rate, SpO2 from EMG, ECG or EEG, alone or in combination to change and often to increase.
Nevertheless, it takes much more effort to obtain physiological data than videos of faces.

Computer vision and supervised learning have come a long way in recent years, redefining the state-of-the-art using deep Convolutional Neural Networks (CNNs).
However, the ability to train deep CNNs for pain assessment is limited by small datasets with labels of patient-reported pain intensities, \emph{i.e.}, annotated datasets such as EmoPain \cite{aung2015automatic}, Shoulder-Pain \cite{lucey2011painful}, BioVid Heat Pain \cite{werner2013towards}.
Particularly, Shoulder-Pain is the only dataset available for visual analysis with per-frame labels.
It contains only 200 videos of 25 patients who suffer from shoulder pain and repeatedly raise their arms and then put them down (onset-apex-offset). 
While all frames are labeled with discrete-valued pain intensities (see Fig. \ref{fig:curve}), 
the dataset is small, the label is discrete and most labels are 0.




Although the small dataset problem prevents us from directly training a deep pain intensity regressor,
we show that fine-tuning from a data-extensive pre-trained domain such as face verification can alleviate this problem.
Our solutions are

\noindent $\bullet$ fine-tuning a well-trained face verification net on additional data with a regularized regression loss and a hidden full-connected layer regularized using dropout,

\noindent $\bullet$ regularizing the regression loss using a center loss,

\noindent $\bullet$ and re-sampling the training data by the population proportion of a certain pain intensity \emph{w.r.t.} the total population.


While our work is not the first attempt of this regularization idea \cite{ding2016facenet2expnet}, to our knowledge we are the first to apply it to the pain expression intensity estimation. Correspondingly, we propose three solutions to address the four issues mentioned above. In summary, the contributions of this work include

\noindent $\bullet$ addressing limited data with expression intensity labels by relating two mappings from the same input face space to different output label space where the identity labels are rich,

\noindent $\bullet$ pushing the pain assessment performance by a large margin,

\noindent $\bullet$ proposing to add center loss regularizer to make the regressed values closer to discrete values,

\noindent $\bullet$ and proposing a more sensible evaluation metric to address the imbalance issue caused by a natural phenomena where most of the time a patient does not express pain.

\section{Related Works}\label{sec:related}
Two pieces of recent work make progress in estimating pain intensity visually using the Shoulder-Pain dataset only: Ordinal Support Vector Regression (OSVR) \cite{Zhao_2016_CVPR} and Recurrent Convolutional Regression (RCR) \cite{zhou2016recurrent}. Notably, RCR \cite{zhou2016recurrent} is trained end-to-end yet achieving sub-optimal performance. Please see reference therein for other existing works.
For facial expression recognition in general, there is a trade-off between method simplicity and performance,
\emph{i.e.}, image-based \cite{ding2016facenet2expnet,fabian2016emotionet} \emph{vs.} video-based \cite{Dapogny_2015_ICCV,liu2014learning,wang2013capturing,guo2012dynamic} methods.
As videos are sequential signals, appearance-based methods including ours cannot model the dynamics given by a temporal model \cite{Dapogny_2015_ICCV} or spatio-temporal models \cite{liu2014learning,wang2013capturing,guo2012dynamic}. 

As regards regularizing deep networks, there exists recent work that regularize deep face recognition nets for expression classification - FaceNet2ExpNet \cite{ding2016facenet2expnet}. During pre-training, they train convolutional layers of the expression net, regularized by the deep face recognition net. In the refining stage, they append fully-connected (FC) layers to the pre-trained convolutional layers and train the whole network jointly.

\section{Regularized deep regressor}
\label{Sec:reg}

\begin{figure}[!t]
    \centering
    \includegraphics[scale=0.75]{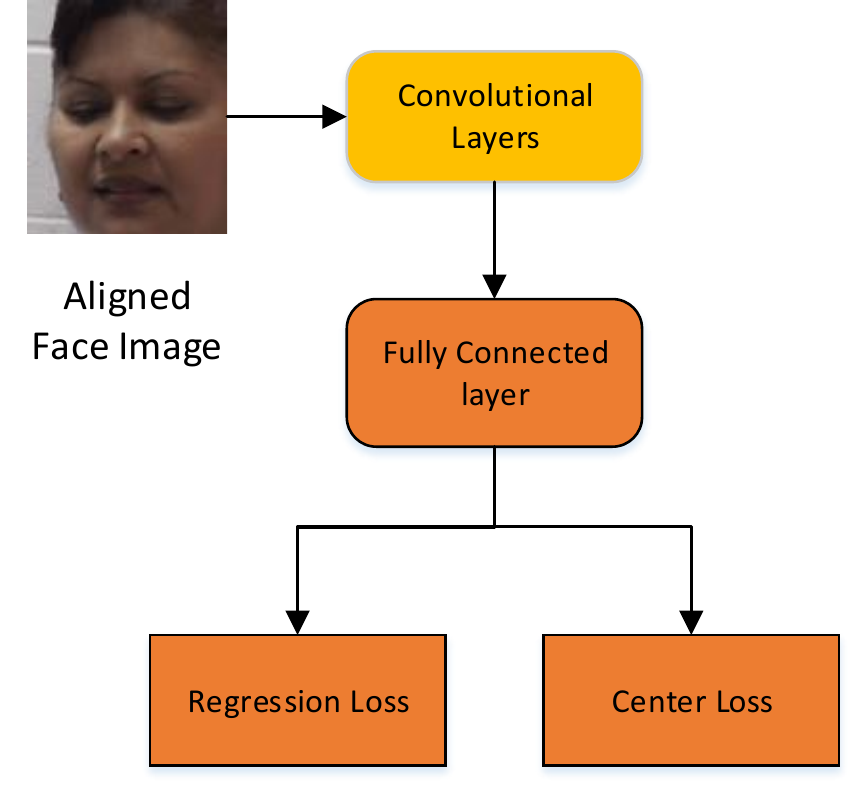}
    \caption{Simplified illustration of the network architecture. The convolution layers are adapted from a state-of-the-art face verification network \cite{wen2016discriminative} while we remove all the fully-connected (FC) layers in \cite{wen2016discriminative} and then add two new FC layers. To avoid over-fitting the limited data, the number of neurons in our hidden FC layer is relatively smaller than the previous layer (50 vs 512), known as Dropout\cite{srivastava2014dropout} as regularization.}
    \label{fig:net1}
\end{figure}

Our network is based on a state-of-the-art face verification network \cite{wen2016discriminative}\footnote{Model available at \url{https://github.com/ydwen/caffe-face}} trained using the CASIA-WebFace dataset contaning $0.5$ million face images with identity labels. 
As a classification network, it employs the Softmax loss regularized with its proposed center loss.
But it is difficult to directly fine-tune the network for pain intensity classification due to limited face images with pain labels.
However, it is feasible to fit the data points $(feature, intensity)$ as a regression problem.
Our fine-tuning network employs a regression loss regularized with the center loss,
as shown in Fig. \ref{fig:net1}.

First, we modify the face verification net's softmax loss to be a Mean Square Error (MSE) loss for regression.
The last layer of such a network is a $\ell 2$ distance layer,
which easily causes gradient exploding due to
large magnitudes of the gradients at initial iterations.
Thus, we replace the MSE loss using
a smooth $\ell 1$ loss with a Huber loss flavor (see Sec. \ref{sec:regloss}).

Secondly, as labels are discrete, 
it is sensible to regularize the loss to make the regressed values to be more ‘discrete’. 
We introduce the center loss \cite{wen2016discriminative} as a regularizer (see Sec. \ref{sec:cls}).

Thirdly, we propose two weighted evaluation metrics in Sec.\ref{sec:wmet} to address label imbalance which may induce trivial method. In the following, we elaborate on the three solutions.
     


\subsection{Regression Loss}\label{sec:regloss}
Similar to conventional regression models, a regression net minimizes the Mean Square Error (MSE) loss defined as
\begin{equation}
    \mathcal{L}_{R_{MSE}} = \frac{1}{N} ( \sigma(\mathbf{w}^T \mathbf{x}) - \tilde{y} )^2
\end{equation}
\label{eqn:regloss}
\vspace{-3mm}

\noindent where $\mathbf{x}$ is the output vector of the hidden FC layer, $\mathbf{w}$ is a vector of real-valued weights, $\tilde{y}$ is the ground-truth label, and
$\sigma(\cdot)$ is a sigmoid activation function $\sigma (x) = \frac{5}{1+e^{-x}}$.
We use $\sigma(\cdot)$ to truncate the output of the second FC layer to be in the range of pain intensity $[0, 5]$.
Here we omitted the bias term for elegance.
The gradient exploding problem often happens due to the relatively large gradient magnitude during initial iterations. 
This phenomenon is also described in \cite{girshick2015fast}. 
To solve this problem, we follow \cite{girshick2015fast} to apply the smooth $\ell 1$ loss 
which makes the gradient smaller than the case with the MSE loss 
when the absolute error $|\sigma(\mathbf{w}^T \mathbf{x}) - \tilde{y}|$ is large.
Different from \cite{girshick2015fast}, our regressor outputs a scalar instead of a vector. 
It is a compromise between squared and absolute error losses:
\begin{equation}
    \mathcal{L}_R = \left\{
                \begin{array}{ll}
                  0.5 |\sigma(\mathbf{w}^T \mathbf{x}) - \tilde{y}|^2, \quad if \quad |\sigma(\mathbf{w}^T \mathbf{x}) - \tilde{y}| < t\\
                  |\sigma(\mathbf{w}^T \mathbf{x}) - \tilde{y}|- t + 0.5 t^2, \quad otherwise\\
                \end{array}
              \right.
\label{eq:smoothl1}
\end{equation}
where $t$ is the turning point of the absolute error between the squared error function and the absolute error function. It has a flavor with the Huber loss. When $t=1$, it works similar with MSE loss since the error is usually below 1. When $t=0$, it is equivalent with the Mean Abosolute Error (MAE) loss. 

\subsection{Regularization Using Center Loss} \label{sec:cls}
\vspace{-2mm}
\begin{figure}[!t]
    \centering
    \includegraphics[scale=0.85]{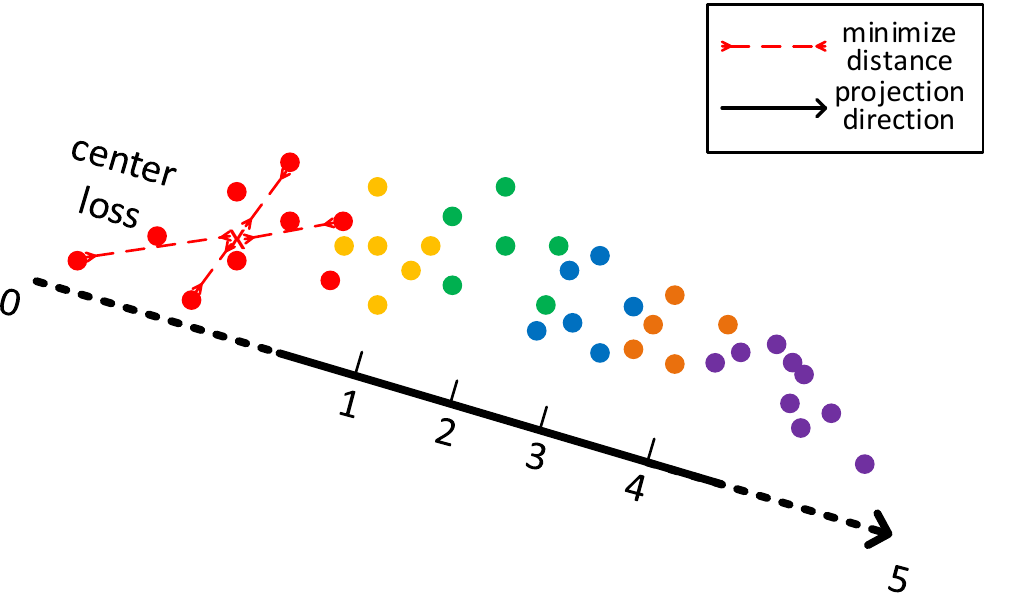}
    \caption{Illustration of how the loss functions works. Each point represents a feature vector in the feature space. By a regression loss, we find a linear projection to project the feature vectors to one-dimension values. The calibration of the coordinate axis is not uniform because we use sigmoid activation, which is not a linear function. Best viewed in color.
    }
    \label{fig:projection}
\end{figure}

Since the pain intensity is labeled as discrete values in the Shoulder-Pain dataset, it is natural to regularize the network to make the regressed values to be `discrete' - during training, to make same-intensity's regressed values as compact as possible (see Fig. \ref{fig:projection}).
We use the center loss \cite{wen2016discriminative} which minimizes the within-class distance and thus is defined as
\begin{equation}
\mathcal{L}_C = \|\mathbf{x} - \mathbf{c}_{\tilde{y}}\|_p^p,
\label{centerloss}
\end{equation}
\noindent where $c_{\tilde{y}}$ represents the center for class $\tilde{y}$ and is essentially the mean of features per class.
$p$ denotes the norm and is typically $\ell 1$ or $\ell 2$. 
We observe from expriments that the center loss shrinks the distances of features that have the same label,
which is illustrated in Fig. \ref{fig:projection}.
To relate it with the literature, it is a similar idea to the Linear Discriminant Analysis yet without minimizing between-class distances. It also has a flavor of the k-means clustering yet in a supervised way.

Now, the center loss is added to the regression loss after the hidden FC layer to induce the loss $\mathcal{L} = \mathcal{L}_R + \lambda \mathcal{L}_C$ where $\lambda$ is a coefficient.
Thus, the supervision of the regularizer is applied to the features.
Different from \cite{wen2016discriminative}, we jointly learn the centers and minimize within-class distances by gradient descent, while \cite{wen2016discriminative}'s centers are learned by moving average.  

\subsection{Weighted Evaluation Metrics}\label{sec:wmet}
\vspace{-2mm}
\noindent Labels in the Shoulder-Pain dataset are highly imbalanced, as 91.35\% of the frames are labeled as pain intensity 0. 
Thus, it is relatively safe to predict the pain intensity to be zero. 

To fairly evaluate the performance, we propose the weighted version of evaluation metrics, \emph{i.e.}, weighted MAE (wMAE) and weighted MSE (wMSE) to address the dataset imbalance issue.
For example, the wMAE is simply the mean of MAE on each pain intensity.
In this way, the MAE is weighted by the population of each pain intensity.

We apply two techniques to sample the training data to make our training set more consistent with the new metrics.
First, we eliminate the redundant frames on the sequences following \cite{Zhao_2016_CVPR}. 
If the intensity remains the same for more than 5 consecutive frames, we choose the first one as the representative frame.
Second, during training, we uniformly sample images from the 6 classes to feed into the network. In this way, what the neural network `see' is a totally balanced dataset.

\section{Experiments} \label{sec:exp}
\vspace{-1mm}
In this section, we present implementations and experiments. 
The project page\footnote{\url{https://github.com/happynear/PainRegression}.} has been set up with programs and data.
\vspace{-2mm}
\subsection{Dataset and Training Details}
\vspace{-2mm}
We test our network on the Shoulder-Pain dataset \cite{lucey2011painful} that contains 200 videos of 25 subjects and is widely used for benchmarking the pain intensity estimation.
The dataset comes with four types of labels. The three annotated online during the video collection are the sensory scale, affective scale and visual analog scale ranging from $0$ (\emph{i.e.}, no pain) to $15$ (\emph{i.e.}, severe pain). 
In addition, observers rated pain intensity (OPI) offline from recorded videos ranging from $0$ (no pain) to $5$ (severe pain).
In the same way as previous works \cite{Zhao_2016_CVPR, zhou2016recurrent, rudovic2013automatic}, we take the same online label and  quantify the original pain intensity in the range of $[0, 15]$ to be in range $[0, 5]$.

The face verification network \cite{wen2016discriminative} is trained on CASIA-WebFace dataset \cite{yi2014learning}, which contains 494,414 training images from 10,575 identities. To be consistent with face verification, we perform the same pre-processing on the images of Shoulder-Pain dataset. To be specific, we leverage MTCNN model \cite{MTCNN} to detect faces and facial landmarks. Then the faces are aligned according to the detected landmarks.

The learning rate is set to $0.0001$ to avoid huge modification on the convolution layers. The network is trained over 5,000 iterations, which is reasonable for the networks to converge observed in a few cross validation folds. We set the weight of the regression loss to be 1 and the weights of softmax loss and center loss to be 1 and 0.01 respectively.

\subsection{Evaluation Using Unweighted Metrics}
\vspace{-2mm}

Cross validation is a conventional way to address over-fitting small dataset.
In our case, we run 25-fold cross validation 25 times on the Shoulder-Pain dataset which contains 25 subjects. 
This setting is exactly the leave-one-subject-out setting in OSVR \cite{Zhao_2016_CVPR} except that OSVR's experiments exclude one subject whose expressions do not have noticeable pain (namely 24-fold).
Each time, the videos of one subject are reserved for testing.
All the other videos are used to train the deep regression network.
The performance is summarized in Table \ref{tab:perf}. It can be concluded that our algorithm performs best or equally best on various evaluation metrics, especially the combination of smooth $\ell 1$ loss and $\ell 1$ center loss. Note that OSVR \cite{Zhao_2016_CVPR} uses hand-crafted features concatenated from landmark points, Gabor wavelet coefficients and LBP + PCA.

\begin{table}[!t]
{\small
\begin{center}
\begin{tabular}{|c|c|c|c|}
\hline
Methods & MAE$\downarrow$ & MSE$\downarrow$ & PCC$\uparrow$ \\
\hline
\hline
smooth $\ell 1$ & 0.416 & 1.060 & 0.524 \\
\hline
$\ell 1$+ $\ell 1$ center loss & {\bf \emph{0.389}} & 0.820 & 0.603 \\
\hline
smooth $\ell 1$ + $\ell 1$ center loss & 0.456 & {\bf \emph{0.804}} & {\bf \emph{0.651}} \\ 
\hline
smooth $\ell 1$ +  $\ell 2$ center loss & 0.435 & 0.816 & 0.625 \\
\hline
\hline
OSVR-$\ell 1$ (\cite{Zhao_2016_CVPR} CVPR'16)  & 1.025 & N/A & 0.600 \\
\hline
OSVR-$\ell 2$ (\cite{Zhao_2016_CVPR} CVPR'16)  & 0.810 & N/A & 0.601 \\
\hline
RCR (\cite{zhou2016recurrent} CVPR'16w)  & N/A & 1.54 & {\bf \emph{0.65}} \\
\hline
\hline
All Zeros (trivial solution) & 0.438 & 1.353 & N/A \\
\hline
\end{tabular}
\end{center}
\vspace{-2mm}
\caption{Performance of our regression network and related works on the Shoulder-Pain dataset 
for the estimation of pain intensity (\emph{i.e.}, pain expression intensity).
MAE is short for mean absolute error deviated from the ground-truth labels over all frames per video.
MSE is mean squared error which measures the curve fitting degree. 
PCC is Pearson correlation coefficient which measures the curve trend similarity ($\uparrow$ indicates the larger, the better).
The best is highlighted in bold.}
\label{tab:perf}
}
\end{table}

\subsection{Evaluation Using Weighted Metrics}\label{sec:imb}
\vspace{-2mm}
In Table \ref{tab:perf}, we provide the performance of predicting all zeros as a baseline. 
Interestingly, on the metrics MAE and MSE, zero prediction performs much better than several state-of-the-art algorithms. 
Now, using the new proposed metrics, the performance is summarized in Table \ref{tab:perf-w}. 
The performance of previous work OSVR \cite{Zhao_2016_CVPR} is no longer below that of predicting all zeros. 
We can also see from Table \ref{tab:perf-w} that the uniform class sampling strategy does help a lot on the new evaluation metrics. Moreover, we have provided the evaluation program in our project page and encourage future works to report their performance with the new evaluation metrics.

\begin{table}[!t]
{\small
\begin{center}
\begin{tabular}{|c|c|c|c|}
\hline
Methods & wMAE$\downarrow$ & wMSE$\downarrow$ \\
\hline
\hline
smooth $\ell 1$ & 1.596 & 4.396 \\
\hline
$\ell 1$ +  $\ell 1$ center loss & 1.388 & 3.438 \\
\hline
smooth $\ell 1$ + $\ell 1$ center loss & 1.289 & 2.880 \\ 
\hline
smooth $\ell 1$ +  $\ell 2$ center loss & 1.324 & 3.075 \\
\hline
$\ell 1$ + $\ell 1$ cente loss + sampling & 1.039 & 1.999 \\ 
\hline
smooth $\ell 1$ + $\ell 1$ center loss + sampling & {\bf \emph{0.991}} & {\bf \emph{1.720}} \\ 
\hline
\hline
OSVR-$\ell 1$ (\cite{Zhao_2016_CVPR} CVPR'16)  & 1.309 & 2.758 \\
\hline
OSVR-$\ell 2$ (\cite{Zhao_2016_CVPR} CVPR'16)  & 1.299 & 2.719 \\
\hline
\hline
All Zeros (trivial solution) & 2.143 & 7.387 \\ 
\hline
\end{tabular}
\end{center}
\vspace{-2mm}
\caption{Performance of our network when evaluated using the weighted MAE and weighted MSE. `sampling' means the uniform class sampling technique is applied. Notably, $\ell 1$ center loss and sampling incrementally boost the performance. }
\label{tab:perf-w}
}
\end{table}
\vspace{-2mm}
\vspace{-2mm}

\section{Summary}
Given the restriction of labeled data which prevents us from directly training a deep pain intensity regressor, fine-tuning from a data-extensive pre-trained domain such as face verification can alleviate the problem.
In this paper, we regularize a face verification network for pain intensity regression. 
In particular, we introduce the Smooth $\ell 1$ Loss to (continuous-valued) pain intensity regression
as well as introduce the center loss as a regularizer to induce concentration on discrete values.
The fine-tuned regularizered network with a regression layer is tested on the UNBC-McMaster Shoulder-Pain dataset and achieves state-of-the-art performance on pain intensity estimation.
The main problem that motivates this work is that expertise is needed to label the pain.
The take-home message is that fine-tuning from a data-extensive pre-trained domain can alleviate small training set problems.

On the other hand, unsupervised learning does not rely on training data.
Indeed, discrete-valued regression is a good test bed for center-based clustering.
Although regularizing a supervised deep network is intuitive, its performance is rather empirical. 
In the future, we need insights about when and why it may function as transfer learning.
Note that no temporal information is modeled in this paper.
As pain is temporal and subjective, prior knowledge about the stimulus needs to be incorporated to help quantify individual differences.

\vspace{-3mm}
\section{ACKNOWLEDGMENT}
\vspace{-3mm}
When performing this work, Xiang Xiang is funded by JHU CS Dept's teaching assistantship, Feng Wang \& Alan Yuille are supported by the Office of Naval Research (ONR N00014-15-1-2356), Feng \& Jian Chen are supported by the National Natural Science Foundation of China (61671125, 61201271), and Feng is also funded by China Scholarship Council (CSC). Xiang is grateful for a fellowship from CSC in previous years.

{
\bibliographystyle{IEEEbib}
\bibliography{icip2017template}
}
\end{document}